\pdfoutput=1
\documentclass[11pt,table]{article}

\usepackage[]{acl}

\usepackage{times}
\usepackage{latexsym}
\usepackage{todonotes}
\usepackage{booktabs}
\usepackage{amssymb}
\usepackage[T1]{fontenc}
\usepackage[utf8]{inputenc}

\usepackage{graphicx}
\graphicspath{ {./figures/} }
\usepackage{microtype}

\newcommand{\probrareclass}{\textsc{Prc}}
\newcommand{\entropy}{\textsc{Entropy}}
\newcommand{\CAL}{\textsc{Cal}}
\newcommand{\coreset}{\textsc{CoreSet}}
\newcommand{\random}{\textsc{Random}}
\newcommand{\iterative}{\textsc{It}}

\newcommand{\cumulative}{\textsc{Cm}}
\newcommand{\pdtb}{\textsc{Ce}}
\newcommand{\thoughtseg}{\textsc{Thought}}
\newcommand{\otherseg}{\textsc{Other}}

\newcommand{\Diss}{$iter_0$}

\newcommand{\consorel}{\textsc{Consonance}}
\newcommand{\dissrel}{\textsc{Dissonance}}
\newcommand{\narel}{\textsc{Neither}}

\title{Transfer and Active Learning for Dissonance Detection:\\ Addressing the Rare-Class Challenge}

\author{
Vasudha Varadarajan$^1$, Swanie Juhng$^1$, Syeda Mahwish$^1$, Xiaoran Liu$^2$\\ {\bf Jonah Luby, Christian Luhmann$^2$ \and H. Andrew Schwartz$^1$ } \\
$^1$Department of Computer Science, Stony Brook University \\
$^2$Department of Psychology, Stony Brook University \\
{\tt \{vvaradarajan,sjuhng,smahwish,has\}@cs.stonybrook.edu} \\
{\tt \{christian.luhmann,xiaoran.liu\}@stonybrook.edu}, 
{\tt jonahluby@gmail.com}
}
\usepackage[table]{}

\begin{document}
\maketitle
\begin{abstract}
%Original: 
%A: 1 sentence to motivate why people should care about fine-tuning and rare class. 
%Although active annotation and learning has been theoretically studied as an improvement over the traditional annotation strategy, there is much to be explored to determine the best practices for practical applications of active learning to NLP. 

While transformer-based systems have enabled greater accuracies with fewer training examples, data acquisition obstacles still persist for rare-class tasks -- when the class label is very infrequent (e.g. < 5\% of samples).
Active learning has in general been proposed to alleviate such challenges, but choice of selection strategy, the criteria by which rare-class examples are chosen, has not been systematically evaluated. 
Further, transformers enable iterative transfer-learning approaches. 
We propose and investigate transfer- and active learning solutions to the rare class problem of dissonance detection through utilizing models trained on closely related tasks and the evaluation of acquisition strategies, including a proposed \textit{probability-of-rare-class} (PRC) approach.
We perform these experiments for a specific rare class problem: collecting language samples of cognitive dissonance from social media.
%We perform these experiments for a specific ``absolute-rarity'' problem: collecting language samples of cognitive dissonance from social media. 
We find that PRC is a simple and effective strategy to guide annotations and ultimately improve model accuracy while transfer-learning in a specific order can improve the cold-start performance of the learner but does not benefit iterations of active learning.
% most active learners face.   %(a) \vasequ{haven't included this in the method yet-- should we go into PDTB vs Kialo pretraining?} 
%"hard problem" is ok if you back it up with how/why it's hard. %In this work, we focus on \sj{subjective expression thus better to eliminate} [a particularly hard problem:] annotating samples of dissonance on social media -- a rare class problem with a natural occurrence rate of 2-5 \% of tweets on Twitter. \sj{better way to put it: we focus on addressing a rare class problem by exploring relevant acquisition strategies, thus improving the proportion of rare-classed samples}  We apply active learning to a classification problem with a rare class, to improve the number of rare class samples in our collected dataset. To do this we explore multiple acquisition strategies relevant to our problem and find that using rare class probability as an acquisition function works just as well as more computationally complex, state-of-the-art acquisition strategies. We also compare two different model update approaches used in literature and find the cumulative update to work better in improving the classification model. 
%\vasequ{how to weave in the cognitive aspect : dissonance - should we include that in the abstract?}
%andy pass 1: 

\end{abstract}

\section{Introduction}

%Cognitive dissonance is a cognitive phenomenon that commonly occurs during everyday decision-making and rationalizing, mostly expressed through language.
% general writing rule: never define a multiword word term by using one of the multiple words (*cognitive* dissonance is a *cognitive* phenomenon...)
% : most people experience it and occasionally express it through language. 

%\bluetodo{Question: should we swap the first two paragraphs? Please check comments }
%\sj{suggestion: I think it's better to move this to after the second paragraph, because I feel our paper emphasizes the importance of studying active learning strategies and transfer learning to solve rare class problem (although cognitive dissonance dataset is equally important)} 
%\va{reply: Since we conduct the study exclusively for cognitive dissonance it helps to motivate it from that angle in my opinion. It was initially the second paragraph but I found it hard to make the narrative flow from "what is cognitive dissonance" to the list of contributions. whoever reads-- please let us know what you think!}
Cognitive dissonance occurs during everyday thinking when one experiences two or more beliefs that are inconsistent in some way~\cite{harmon2007cognitive}. 
%A: why should we care about dissonance; must establish first:
Often expressed in language, dissonance plays a role in many aspects of life, for example  affecting health-related behavior such as smoking~\cite{chapman1993self} and contributing to the development of (and exit from) extremism~\cite{dalgaard2013promoting}. %\textcolor{green}{Something about how people's desire to be self-consistent makes manifestations of dissonance rare?} 
However, while the phenomenon is common enough to occur on a daily basis, dissonance is still relatively rare among the myriad of other relationships between beliefs that occur across random selections of expressions and thus makes the automatic detection of it a rare-class problem. 
%, possibly due to the human tendency to be consistent in one's beliefs because of judgment from expressing inconsistent beliefs \citep{tedeschi1971cognitive}. 
%likely due to the tendency for one to try to keep their own beliefs consistency \citep{tedeschi1971cognitive}.how about: 
%This may be attributed  In other words, disclosing dissonance within oneself may lead to loss of credibility.

% \sj{On the other hand, this paper \citep{pyszczynski1993} says that ``Nisbett and Wilson (1977) reviewed the dissonance literature for evidence that subjects exhibit awareness of any of the processes posited by the etheory and found little evidence of such awareness. Nisbett and Wilson argued that this lack of awareness is consistent with their general thesis that subjects rarely have direct access to the cognitive processes through which dissonance and other related effects.'' This could explain why expression of dissonance is hard to find in social media language but in a different way}

%Although language has been studied in the NLP community as structured constructions of words or phrases \cite{naseem2021comprehensive} or their semantic roles \cite{li2022survey}, language is fundamentally a cognitive phenomenon, 
%\sj{thus it is important to study language in the cognitive aspect. (???????)}   \va{
%how about: 
%thus meriting the study of cognitive aspects of language as well.
%}
% [there is merit to obtaining expressions of cognitive phenomena directly.]
Despite recent advances in modeling sequences of words, 
 rare-class tasks -- when the class label is very infrequent (e.g. < 5\% of samples) -- remain
challenging due to the low rate of positive examples. 
Not only are more random examples necessary to reach a substantial amount (e.g. 1,000 examples to reach just 50 examples of the rare class) but also it is easy for human annotators to miss the rare instances where dissonance is present. 
Here, we develop and address the challenges of creating a resource for language-based assessment of dissonance.

%Most of the extant works do not focus on real-time, large-scale active annotation for a novel classification problem as annotating cognitive dissonance on social media. 
\begin{figure}[!t]
    \centering
    \includegraphics[width=7.2cm]{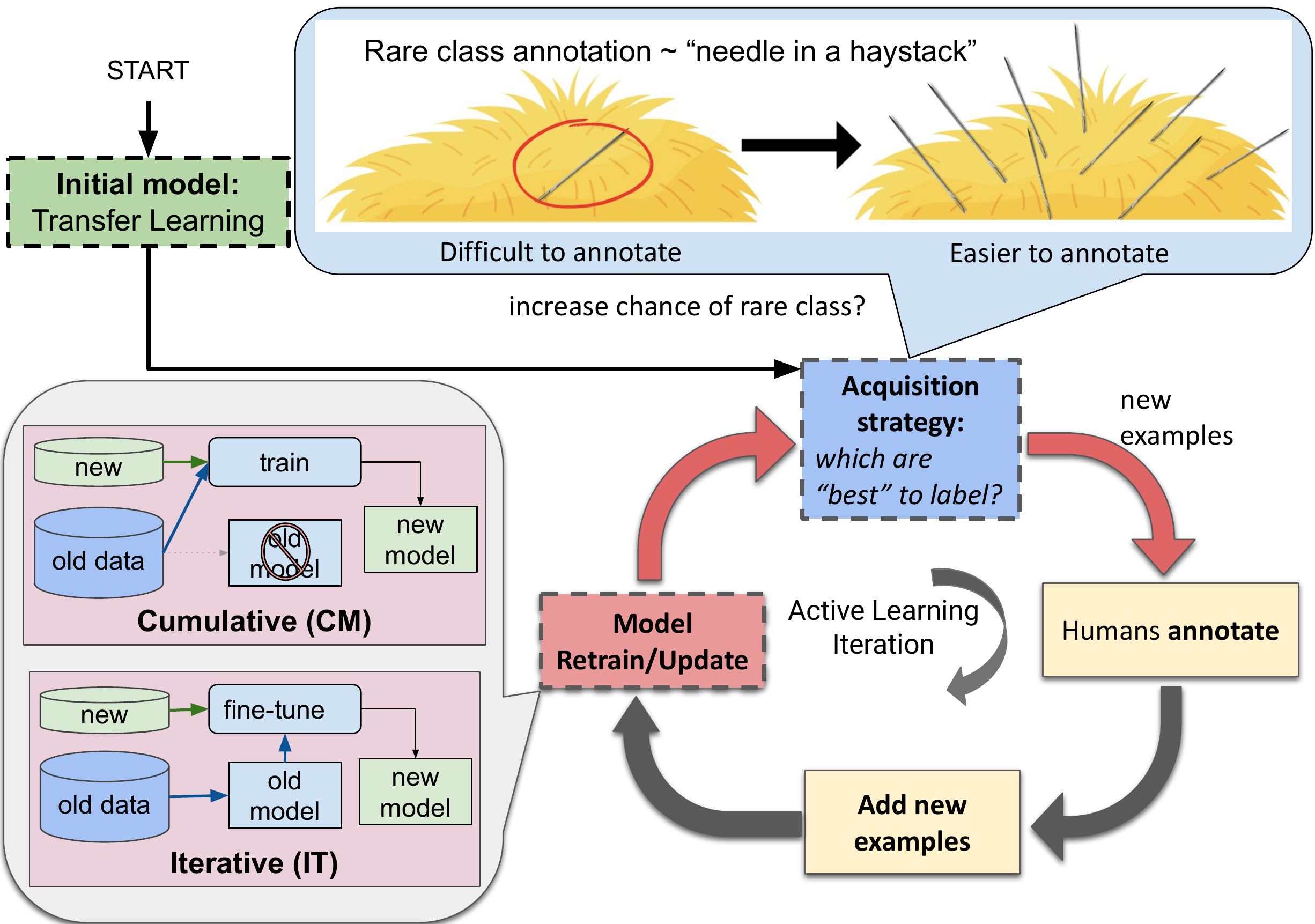} % link to drawing: https://docs.google.com/drawings/d/1whza7Hmtv16NnDHjdIAVJuMDsOVDkQ_OHqo4ADasIKM/edit?usp=sharing
    \caption{
    Demonstration of the active learning (AL) loop in general. Our paper examines the three highlighted steps: (i) Bootstrapping with TL model, (ii) Acquisition strategy, and (iii) Model update.   
    % Three steps in the AL loop are studied: (i) Bootstrapping with TL model, (ii) Acquisition strategy and (iii) Model update. 
    % We study combinations of two similar tasks as candidate models for transfer learning to bootstrap the AL loop.
    % Multiple acquisition strategies are studied in the context of increasing the proportion of rare samples in the annotation sets in each round. 
    % Model update can be done in two ways as shown in the figure: the difference between cumulative and iterative approach for active learning is evident in the model updating step where iterative approach progressively finetunes a model using newly acquired data whereas the cumulative approach takes a base model and trains it from scratch with all of the data collected so far.
    }
   
    \label{fig:AL_loop}
\end{figure}

%A: not sure what this means? larger language models makes less annotation necessary because they need fewer examples. 
%The emergence of new tasks and larger language models % with increasing number of parameters in NLP 
%has far exceeded the speed at which data annotations could be done for supervised tasks.

Active learning using large language models presents both new opportunities and challenges. 
%A: not sure this is true; LLMs are intensive to train from scratch but not afterward; also would definitely need a reference from the last couple yearschallenging due to the models' poor calibration, resulting in rarely being uncertain during inference and requirement of a relatively large amount of training data from the outset \cite{guo2017calibration}.
On the one hand, LLMs offer unmatched representations of documentations, able to achieve state-of-the-art language understanding task performance with transfer learning, often only with a few iterations of fine-tuning~\cite{liu2019roberta}.
On the other hand, representations are high-dimensional, and models trained or fine-tuned with only a small number of examples are prone to overfitting, especially when there is a large class imbalance as in rare-class problems.
While LLMs have enabled attempting more and more complex semantic challenges across a growing list of tasks, getting annotated examples for such problems can become a bottleneck due to its time- and labor-intensiveness~\cite{survey_HITL}.
Since data-centric improvements for more novel tasks can provide a faster path than model-centric improvements~\cite{ng2021mlops}, active learning can be a way forward to be both data-centric and address bottlenecks in label acquisition -- it aims to reduce annotation costs as well as alleviate the training data deficiency that large language models face. 
% Active learning for large language models can provide a solution, but
%In addition

However, while active learning has been studied for multiple natural language tasks \citep{shen2017deep, liang2019new}, little is known about active learning acquisition strategies for LM-based approaches, especially for rare-class problems. %experimented with acquisition strategies used to pick out new samples for annotation in each round. 
\textit{High data imbalance} coupled with \textit{very less training data} poses the challenge of ``absolute rarity'' \cite{al2016transfer}, as in our task of dissonance detection. 
We address this problem by using a novel combination of evaluating the ordering of transfer learning from similar tasks to cold-start the active learning loop, and by acquiring with a relatively simple %but novel 
acquisition strategy focused on \textit{probability-of-rare-class} (PRC) to increase the rare class samples.

Our contributions include: 
(1) finding that bootstrapping AL models with transfer learning on closely related tasks significantly improves rare class detection; 
(2) a novel systematic comparison of five common acquisition strategies for active learning for a rare class problem; %,  finding \probrareclass{} performs the best in terms of improving the model; 
(3) a systematic comparison of two different approaches to handling AL iterations for LLMs -- cumulative and iterative fine-tuned model updates finding the cumulative approach works best; (4) evaluating annotation costs of a rare-class task, finding that minimum annotation cost does not necessarily lead to better models, especially in realistic scenarios such as \textit{absolute rarity}; and (5) release of a novel dataset for the task of identifying cognitive dissonance in social media documents.

%Active learning strategies are many but not  a lot has been explore for difficult annotations / rare class as well (cite the VQA paper) -  https://arxiv.org/abs/2107.02331

%dissonance is a rare class problem

\section{Related Work}
\label{sec:related}
%\sj{Typical ACL papers do not divide related work into sections. Should we get rid of such divisions?} \va{Does it sound too choppy while reading? Was finding it hard to make one cohesive section}

Active learning in NLP has been largely studied as a theoretical improvement over traditional ML %in the context of acquisition strategies 
for scarce data. 
In this work, we specifically investigate \textit{pool-based} active learning, or picking out samples to annotate from a larger pool of unlabeled data, and particularly data for a \textit{rare-class} problem where LMs are not well-understood yet.

\begin{table}[!ht]
\begin{small}
\centering
\hskip-0.4cm
\begin{tabular}{llllll}
\toprule
         & F1-macro & F1-Dis & Prec-Dis   & Rec-Dis & AUC     \\
\midrule
Diss alone & 0.478 & 0.000  & 0.000 & 0.000  & 0.500 \\
\midrule
%\multicolumn{4}{c}{Transfer-Learning Alone}\\
%\midrule
%Debate  & 0.??? & 0.269 & 0.287  & \textbf{0.252}  & 0.595 \\
Debate  & \textbf{0.595} & \textbf{0.319} & 0.349  & \textbf{0.278}  & \textbf{0.620} \\
\pdtb{}    & 0.487 & 0.210 & \textbf{0.558} & 0.129 & 0.602 \\
%\pdtb{}    & 0.??? & 0.237 & 0.545 & 0.152 & 0.597 \\
Deb; \pdtb{} & 0.540 & 0.211 & 0.349 & 0.152 & 0.583 \\
%Deb; \pdtb{} & 0.??? & 0.235 & 0.353 & 0.176 & 0.582 \\
%Deb→\pdtb{} & 0.297 & 0.145 & 0.195 & 0.548          \\
%\pdtb{}→Deb & \textbf{0.733} & 0.169  & \textbf{0.274} & \textbf{0.659}\\
%Deb; \pdtb{}; Diss & 0.353 & 0.176 & 0.235 & 0.582          \\
%\midrule
%\multicolumn{4}{c}{Transfer and Continue Training}\\
%\midrule

%Deb $\rightarrow$ \pdtb{} $\rightarrow$ Diss & 0.???  & 0.??? & 0.???  & 0.??? \\
%\pdtb{} $\rightarrow$ Deb $\rightarrow$ Diss & 0.???  & 0.??? & 0.???  & 0.??? \\

\bottomrule
\end{tabular}

\caption{
Performance of models pretrained on two similar tasks, separately and combined, based on development and test set from $iter_0$. Precision and Recall reported for Dissonance class. ``$;$'' refers to combining the two datasets. \textbf{Bold} represents best in column. Training with dissonance dataset alone doesn't help the model-- this shows the usefulness of transfer learning to cold-start active learning, especially on transfer from Debate. }

% This table shows how pretraining on two similar tasks or a combination of them already helps with the dissonance detection task, without any initial data to train on.
% here
% ``$;$'' refers to combining the two datasets. The F1 and AUC of the Deb model is better than the PDTB model when directly tested on dissonance data, while training on the dissonance training set doesn't perform better than chance (AUC=0.5), showing the usefulness of transfer learning to cold-start active learning, especially on transfer from Debate.

%Training for a specific task for big data results in overfitting and poor performance on the task of interest, as can be seen in the best performance with using 25\% of the datasets. 
 %}
\label{tab:kialo_pdtb}
\end{small}
\end{table}

\begin{table}[!ht]

\begin{small}
\centering
\hskip-0.5cm
\begin{tabular}{p{0.25\linewidth}p{0.1\linewidth}p{0.1\linewidth}p{0.08\linewidth}p{0.08\linewidth}l}
\toprule
         & F1-macro & F1-Dis  & Prec-Dis & Rec-Dis & AUC     \\
\midrule

\multicolumn{6}{c}{Transfer-Learning Alone}\\
\midrule
%Deb; PDTB  &0.???& 0.235 & 0.353 & 0.176 & 0.582          \\
%Deb→PDTB  &0.???& 0.195 & 0.297 & 0.145 & 0.548          \\
%PDTB→Deb  &0.???& \textbf{0.274} & \textbf{0.733} & 0.169  & \textbf{0.659}\\
Deb; \pdtb{} & 0.520 & 0.212 & 0.442 & 0.140 & 0.593 \\
Deb→\pdtb{}  &0.495& 0.170 & 0.349 & 0.112 & 0.544          \\
\pdtb{}→Deb  &0.487& 0.243 & 0.744 & 0.146  & \textbf{0.666} \\
\midrule
\multicolumn{6}{c}{Transfer and Continue Training}\\
\midrule
Deb; \pdtb{}; \Diss{} & 0.458 & 0.033 & 0.100 & 0.02  & 0.507          \\
Deb$\rightarrow$\Diss{} & 0.564 & 0.296& 0.236 & 0.400 & 0.554 \\
\multicolumn{1}{l}{Deb$\rightarrow$\pdtb{}$\rightarrow$\Diss{}} &0.532 & 0.143  & 0.146 & 0.140  & 0.531 \\
\pdtb{}$\rightarrow$Deb$\rightarrow$\Diss{} & 0.585& 0.229  & 0.296 & 0.186  & \textbf{0.572} \\

\bottomrule
\end{tabular}

\caption{
The zero-shot performance of models further fine-tuned from those in Table \ref{tab:kialo_pdtb}. ``$;$'' refers to combining the two datasets, ``→'' indicates iteratively fine-tuning on each task, and \textbf{bold} represents the best in column. Scores based on development and test set from $iter_0$.
The order matters for fine-tuning: we find the  \pdtb{}$\rightarrow$Deb performs the best in zero-shot setting. 
}
% This table explores if further finetuning the transfer learning models from Table \ref{tab:kialo_pdtb} improves the zero-shot performance for the dissonance task. We find that not only finetuning helps, but the order of pretraining on the two tasks matters when it comes to transfer learning for the dissonance task. ``$;$'' refers to combining the two datasets and ``→'' refers to iteratively finetuning on each task. We find the  PDTB$\rightarrow$Deb performs the best in zero-shot setting. 
% here
% The F1 and AUC of the PDTB→Deb model is comparable to the models finetuned to the newly annotated dissonant data in Table \ref{tab:comparison-AL}, which shows the incredible zero-shot capacity of the transfer learning models. On further training with the dissonance dataset, we find that the performance decreases-- but this is attributed to the small size of the $iter_0$ data and the effect of heterogenous domain transfer -- containing only 43 samples of cognitive dissonance.

%theshows how pretraining on two similar tasks or a combination of them already helps with the dissonance detection task, without any initial data to train on.
%Training for a specific task for big data results in overfitting and poor performance on the task of interest, as can be seen in the best performance with using 25\% of the datasets. 
%The scores are the AUC for a binary classification (dissonance or not) for initial dissonance dataset v1.

\label{tab:kialo_pdtb_finetuning}
\end{small}
\end{table}

\paragraph{Acquisition strategies}Sampling strategies for active learning can be broadly classified into three: uncertainty sampling  \citep{shannon1948mathematical, Wang2014ANA, marginentropy2011}, representative (or diversity) sampling \citep{cluster_margin, sener2018active, gissin2019discriminative}, and the combination of the two \cite{zhan2022comparative}.
The uncertainty sampling strategies that employ classification probabilities, Bayesian methods such as variational ratios \citep{freeman1965elementary}, and deep-learning specific methods \citep{houlsby2011bayesian} often use epistemic (or model) uncertainty. We choose maximum entropy to represent the uncertainty sampling, since it is usually on par with more elaborated counterparts \cite{tsvigun-etal-2022-towards}. As a popular diversity sampling baseline to compare against, we pick select CoreSet \citep{sener2018active}.
The state-of-the-art methods combine these two strategies in novel ways, such as using statistical uncertainty in combination with some form of data clustering for diversity sampling \citep{cartography, badge}. Our work uses Contrastive Active Learning \citep{margatina-etal-2021-active} to represent this strategy.
%We pick out five different AL strategies for comparison based on <deep AL paper> for our work, but we also considered other state-of-the-art approaches \cite{zhang-plank-2021-cartography-active, yuan-etal-2020-cold} as candidate strategies, the reasoning explained in \ref{subsec:al-strategies}.

On the other hand, \citet{karamcheti-etal-2021-mind} and \citet{Munjal_2022_CVPR} claim there is rather small to no advantage in using active learning strategies, because a number of samples might be collectively outliers, and existing strategies contribute little to discover them and instead harm the performance of subsequent models. 
Researchers recently have also focused on the futility of complex acquisition functions applied to difficult problems and argued that random acquisition performs competitive to more sophisticated strategies, especially when the labeled pool has grown larger \citep{sener2018active, ducoffe2018adversarial}. Furthermore, a large-scale annotation of randomly sampled data could be less expensive than sampling data in each round of active learning.

\paragraph{Cold-Start AL} While the problem of cold-start exists in acquiring samples through active learning, some work has been done to combat this by leveraging the learned weights in pretrained models \cite{yuan-etal-2020-cold}. However, there is much to gain from the field of transfer learning especially for rare class problems, as seen in \citet{al2016transfer}. We borrow the concept of heterogeneous transfer learning \cite{day2017survey, comprehensiveTL} and transfer the model weights directly obtained from pretraining on closely related (but different) tasks on completely different domains. This helps 
%\sj{do you mean models?} [tasks]
models to improve the zero-shot ability for rare class detection. Such methods have been explored in traditional machine learning \citep{kale2013accelerating} but not in the 
%\sj{what does this mean?} 
era of large language models to the best of our knowledge.

\paragraph{Rare class AL}There has been a growing number of applications of active learning in data imbalance and rare class problems. Such works include \citep{similar_neurips, Choi_2021_CVPR} which proposed frameworks to improve model performance with data imbalance but failed to check the feasibility and costs in a real-world, active annotation setting where not only is rare class very infrequent ($~4\%$) but very few ($< 70$) 
%\sj{QUESTION: out of how many? better to mention percentage?} \va{I was trying to mention how the problem is not just low percentage (that would be just "rarity"), but the total number of rare class examples is just so low for LLM to learn from -- so "absolute rarity"-- reworded by including both the number and percentage} 
examples of the rare class exist due to small dataset size (``absolute rarity''). They also fail to compare against a simple, rare class probability of the model. Many studies also focus on rare class \textit{discovery}, or finding outlying samples that do not fall under the existing categories % for a domain
\citep{rare_class_disco_2011, haines2014active}. This is different from our task which focuses on the \textit{detection} of a rare class.

% \paragraph{Cognitive Dissonance}
\section{Task}
\label{subsec:cogdis}
Cognitive dissonance is a phenomenon that happens when two elements of cognition (i.e. thoughts, experiences, actions, beliefs) within a person do not follow one another or are contradictory, and consonance is when one belief follows from the other \cite{harmon2019introduction}. 
Cognitive dissonance raises psychological discomfort, encouraging a person to resolve the dissonance. 
%The discomfort occurs when one element of knowledge does not align with a person's existing values or beliefs. 
As the magnitude of dissonance increases, the pressure to resolve it grows as well \cite{harmon2008left, mcgrath2017dealing}. 

Social psychology has used this human tendency to resolve dissonance to understand important psychological processes such as determinants of attitudes and beliefs, consequences of decisions, internalization of values, and the effects of disagreement among persons \cite{harmon2019introduction}. 
Dissonance is also related to anxiety disorders, relevant to understanding extremism and predicting cognitive styles of users. %\bluetodo{Syeda/Vasudha will add a description of cognitive dissonance}
Our approach to annotating cognitive dissonance on social media is motivated by the two-stage annotation approach described in \citet{varadarajan2022disso}. 
To the best of our knowledge this is the first social media dataset for cognitive dissonance.

\section{Methods}

\subsection{Annotation and Dataset}
\label{subsec:anno_dataset}
%The Theory of Cognitive Dissonance proposed by Leon Festinger in 1957 explained that dissonance begins when two elements of knowledge (cognition) are relevant or irrelevant to one another. Once two cognitions are established to be relevant to one another, it is understood as either consonant or dissonance. Pairs of cognitions are consonant when one follows from another, but if one cognition is obverse to the other, it is determined to be dissonance \cite{harmon2019introduction}. Cognitive dissonance raises psychological discomfort, encouraging a person to resolve the dissonance. The discomfort occurs when one element of knowledge does not align with a person's existing values or beliefs. Therefore, as the magnitude of dissonance increases, the pressure to reduce dissonance increases \cite{harmon2008left}. Social psychology has used this effect to resolve dissonance to understand important psychological processes such as determinants of attitudes and beliefs, consequences of decisions, the internalization of values, the effects of disagreement among persons, and more \cite{harmon2019introduction}.
 
\begin{figure}
    \centering
    \includegraphics[width=7.5cm]{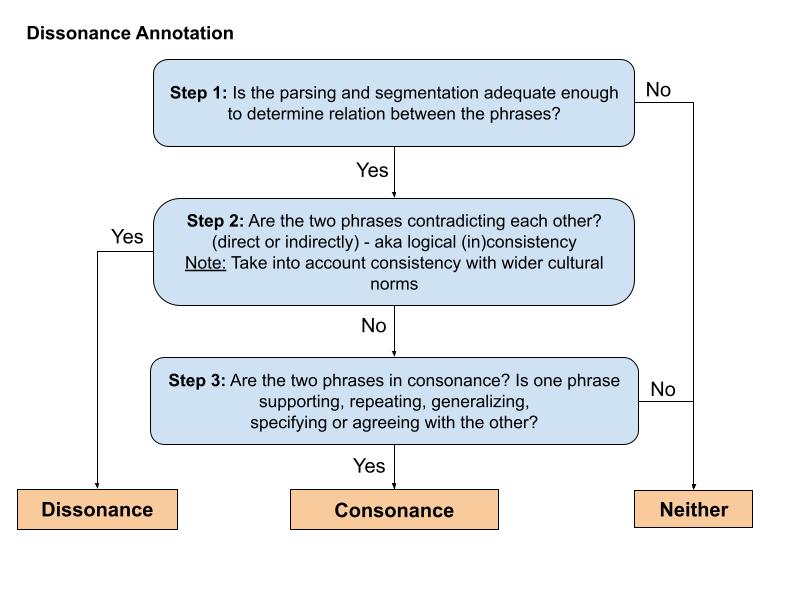} % link to drawing: https://docs.google.com/drawings/d/1whza7Hmtv16NnDHjdIAVJuMDsOVDkQ_OHqo4ADasIKM/edit?usp=sharing
    \includegraphics[width=7.5cm]{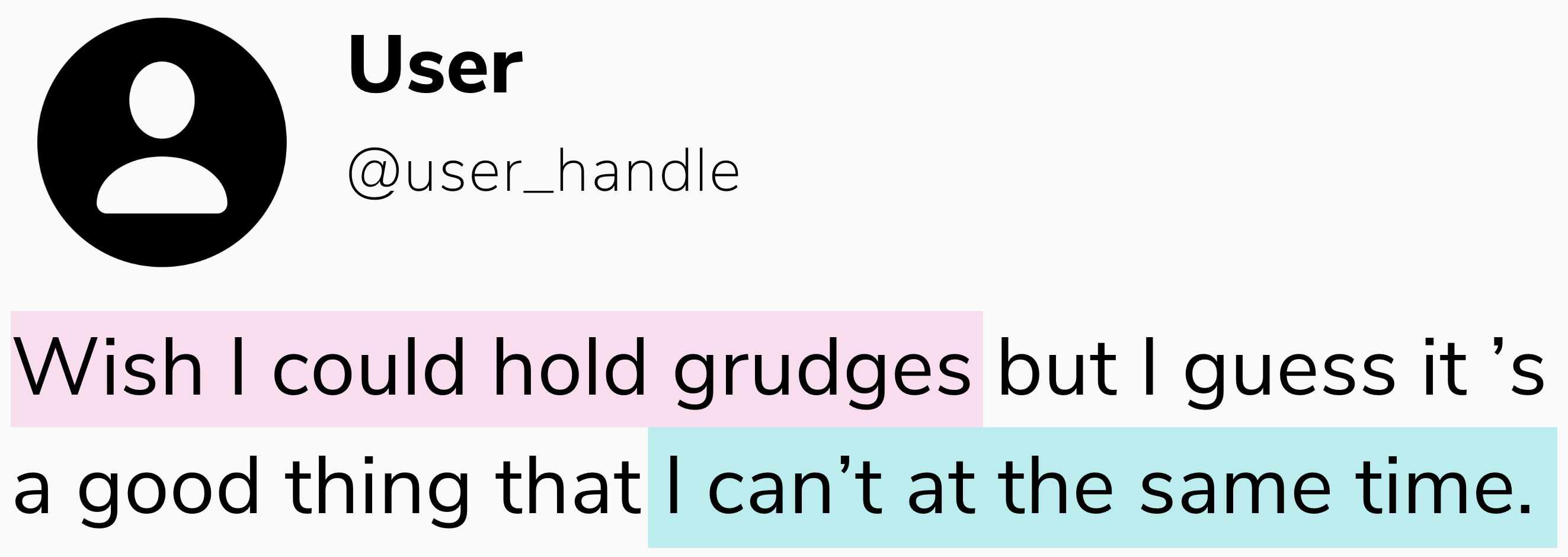}
    \caption{Above: Flowchart describing the steps for the annotators to label tweets as \dissrel{}, \consorel{}, or \narel{}. \\Below: An example of a pair of \thoughtseg{} segments in a tweet annotated as dissonance. }
   
    \label{fig:dissonance_instructions}
\end{figure}

%\begin{figure}[!ht]
%    \centering
%    \includegraphics[width=7.5cm]{figures/dissonance_example.png} % link to drawing: https://docs.google.com/drawings/d/1whza7Hmtv16NnDHjdIAVJuMDsOVDkQ_OHqo4ADasIKM/edit?usp=sharing
%    \caption{Example of a discourse pair annotated as dissonance.}
   
 %   \label{fig:dissonance_instructions}
%\end{figure}
 Following the definition of cognitive dissonance (see \S\ref{subsec:cogdis}), we treat discourse units as semantic elements that can represent beliefs. 
 A discourse unit consists of words or phrases that have a meaning \cite{polanyi1988formal}, and then cognitive dissonance is analogous to a discourse relation between two discourse units.  Recent work \cite{discre2022}, points to understanding discourse relations as representations in a continuous vector space, motivating us to look at cognitive dissonance, too, as a relationship between two ``thought'' discourse units. 
 %or between a ``thought'' and an ``action'' discourse unit. 
% We henceforth use the word $belief$ to refer to a discourse unit that is either a ``thought'' or an ``action'' (i.e. potential elements to be in dissonance).
 %Discourse parsing is a task that chunks a larger piece of text into multiple coherent discourse units. 

%To this end, 
We build a dissonance dataset by first sampling posts between 2011 and 2020 on Twitter. The tweets were parsed into discourse units using the parser by \citet{wang-etal-2018-toward} which uses the PDTB framework.
\footnote{PDTB \cite{prasad2008penn} and RST \cite{mann1987rhetorical} are the two major frameworks for discourse parsing; we use the former for this work since PDTB is lexically grounded and identifies discourse relations using lexical cues. 
While the RST framework could be helpful since rhetorical relations are viewed as cognitive entities \cite{taboada2006rhetorical}, the complex relationships defined with RST's nested structures can complicate our search for cognitive dissonance samples at the preliminary stage of data collection.}
%The annotation is carried out in two stages. 

Each discourse unit in a document is initially annotated into \thoughtseg{} 
%\actionseg{}, 
or \otherseg{}. \footnote{This was a simpler, large-scale annotation to pick out discourse units describing author's own beliefs. We do not go into the details of this specific annotation since it is not pertinent to this work.}
%using a RoBERTa-based classifier trained on a small dataset of 1,145 annotated tweets parsed into 4,137 discourse segments -- 3,120 \thoughtseg{}, 147 \actionseg{} and 870 \otherseg{}. Both \thoughtseg{} and \actionseg{} convey the tweet author's own beliefs.
A \thoughtseg{} is a discourse unit describing the author's own beliefs, experiences and actions and are potential elements to be in dissonance. \otherseg{} comprises of anything else, from  meaningless phrases to coherent beliefs that belong to someone other than the author. % -- these are not considered for the annotation.
%\bluetodo{Explain the action-thought classifier in short paragraph }
%For the Thought-Action classifier- First, each unit is annotated as \thoughtseg{} or \otherseg{}. A \thoughtseg{} constitutes of all forms of knowing and awareness: a fact, claim, or statement is a thought. 
%Anything not considered to be a \thoughtseg{} is marked as an \otherseg{}. 
For the annotation of dissonance, pairs of \thoughtseg{} units 
%or pairs of a \thoughtseg{} and an \actionseg{} 
from each tweet are extracted, and then annotated to compose \consorel{}, \dissrel{} or \narel{} according to the framework described in Figure \ref{fig:dissonance_instructions} -- a three-class annotation. This framework was developed from annotator training to spot examples of dissonance, followed by discussion with a cognitive scientist. 
 %We henceforth use the word $belief$ to refer to a discourse unit that is either a \thoughtseg{} or an \actionseg{} (i.e. potential elements to be in dissonance).
 
%However, the active learning strategies use models that are binary classifiers, i.e. dissonance vs. not dissonance, since \sj{could sound ambiguous to others?} [the other two classes] are abundant and are easily captured by the models with >90\% accuracy, and because we aim to specifically focus on improving dissonance detection. %The v1 validation set, consisting of 1714 discourse unit pairs annotated as \consorel{}, \dissrel{} or \narel{} was used to pick out the best model to use at the bootstrapping step and was annotated traditionally (i.e. not involving active learning). 
Among a random selection of tweets, the natural frequency of the \dissrel{} class is around 4\%. 
%\bluetodo{include the number of dissonance added in each iter or show in the difficulty table}
The annotations were carried out by a team of three annotators, with the third annotator tiebreaking the samples disagreed by the first two annotators. 

\paragraph{Initial set ($iter_{0}$)}  This dataset is used to select the best transfer model to effectively cold-start the AL loop. We start with a total of 1,901 examples of dissonance task annotations, which we split into a training set of 901 examples (henceforth, $iter_{0}$) with 43 examples of dissonance (4.77\%) picked randomly from discourse-parsed tweets. We create initial development and test sets with 500 examples each. They were created such that all the \thoughtseg{} pairs that were a part of a single tweet belong to the same set. 

\paragraph{Final development and test datasets} We gather additional 984
%\sj{do you mean tweets?}\va{no, I meant annotated discourse unit pairs -- we can call them examples if it is better? I actually wanted to point out that further annotations were done separately from the experiments described in this paper just to create dev and test set; but at the same time not bring too much attention to how it was done  (we used a mix of acquisition strategies) because it's only used for testing and finding the best model} 
annotations for development set and 956 annotations for test set in addition to the previously mentioned 500 for each, summing up to 1,484 development examples ($dev$) and 1,456 test examples ($test$) with around 10\% dissonance examples in each, to account for increased frequency of occurrence of the rare class after incorporating novel acquisition strategies.

\subsection{Modeling}
%\bluetodo{ 1. iterative vs cumulative in one table (average the iter 1 and 2), 2. choice of acquisition strategies , 3. difficulty and timing (+  include IAA)in one table, same order as table 1, grab the std dev for timing and diff, } 
\subsubsection{Architecture}
\label{subsub:arch}

A RoBERTa-based dissonance classifier is used consistently across all the experiments in this paper: for any two \thoughtseg{} segments belonging to a single post, the input is in the form of ``$\textrm{[CLS] } segment_1 \textrm{ [SEP] } segment_2 \textrm{ [SEP]}$''. We take the contextualized word embedding $\mathbf{x} \in \mathbb{R}^d$ of $\textrm{[CLS]}$ in the final layer and feed it into the linear classifier: $y = \textrm{softmax}(W\mathbf{x} + \mathbf{b})$,
where $W \in \mathbb{R}^{d \times 2}, \mathbf{b} \in \mathbb{R}^{2}$ is 
a learned parameter. We trained the model parameters with cross entropy 
loss for 10 epochs, using AdamW optimizer with the learning rate of 3 × 
10\textsuperscript{-5}, batch size of 16, and warm up ratio of 0.1. To 
avoid overfitting, we use early stopping (patience of 4) with the AUC score.
%For each AL strategy and model update approach, we conduct two rounds of active annotation and learning loop, on the transfer model \pdtb{}$\rightarrow$Deb$\rightarrow$\Diss{} from Table \ref{tab:kialo_pdtb_finetuning} starting with $iter_0$ trained on a RoBERTa-based \sj{dissonance classifier?} model, input in the form of ``$\textrm{[CLS] } du_1 \textrm{ [SEP] } du_2 \textrm{ [SEP]}$''. 
We run the AL experiments on the datasets delineated in \S\ref{subsec:anno_dataset}.
%\footnote{we repeated the annotation process with a held out test and dev set to be able to systematically compare the methods in this work.}

%based on the criteria in Figure \ref{fig:dissonance_instructions}. 
%The initial base model is referred to as iter 0, and the results are reported in Table \ref{tab:comparison-AL}.
%\sj{I feel this paragraph is redundant} 
While the annotations are for three classes (Figure \ref{fig:dissonance_instructions}), the models used for AL across all strategies classify labels to binary level (dissonance or not dissonance), as we are focused specifically on the dissonance class 
% \sj{I feel this sentence should be placed elsewhere} \va{could you suggest a location/section?} 
-- while dissonance is rare, it is also essential to perform well in detecting this class.
% since it is the focus of our study. 

\subsubsection{Bootstrap with Transfer Learning}
%\sj{need to enhance readability of this paragraph by changing the font of key concepts \& models, splitting the paragraph into several, etc.}
We explore cold-starting the active annotation process using a transfer of model weights trained on similar tasks.  
\paragraph{PDTB-Comparison/Expansion (\pdtb{})} 
The PDTB framework defines discourse relations at three hierarchies: Classes, Types and Subtypes. Of the four classes viz. Temporal, Contingency, Comparison, and Expansion,
the Comparison class ``indicates that a discourse relation is established between two discourse units in order to highlight prominent differences between the two situations'' \cite{prasad2008penn}. While this class is different from \dissrel{}, it is useful in capturing discord between the semantics of two discourse units. The Expansion class is defined to ``cover those relations which expand the discourse and move its narrative or exposition forward,'' which is closer to our conception of \consorel{}.
We thus identify one similar task to be classifying discourse relations as Comparison or Expansion (\pdtb{}).
The \pdtb{} dataset consists of 8,394 (35.12\%) Comparison class examples and 15,506 (64.88\%) Expansion class examples. The  model  was trained on the same architecture as explained in  \S\ref{subsub:arch} with $segment_1$ as the first discourse unit (Arg1) and $segment_2$ as the second discourse unit (Arg2) and the output indicating Comparison or Expansion class. For the training, 10\% was set aside as the development set to pick the best performing model on the \pdtb{} task.

\paragraph{Dissonant Stance Detection (Debate)} The dissonant stance detection task classifies two statements in a debate to be in agreement (consonant stance) or disagreement (dissonant stance) independent of the topic that is being debated upon as described in \citet{varadarajan2022disso}. Dissonant stance is different from \dissrel{} in two ways: (a) each input segment is a complete post consisting of multiple sentences arguing for a stance/topic whereas in our task, they are discourse units; and (b) while both are social media domains, our task uses a more personal, informal language while debate forums use impersonal language citing facts, not author's subjective beliefs. But the tasks are similar in the detection of dissonance between two segments, and we identify it as a potential task to transfer learn from. The statements were extracted from a debate forum consisting of 34 topics with 700 examples each (total 23,800 samples). There were 8,289 dissonant stance examples (34.82\%) in the dataset. While the dataset has three labels -- consonant stance, dissonant stance and neither --, we train a binary classifier on top of the RoBERTa layers to detect dissonant stance or not dissonant stance, keeping the task similar to the model we use in the AL iterations. 

Both of these tasks involve two statements/phrases/sentences as inputs, and the output is Comparison/Expansion in the first case, and Dissonant/Not Dissonant stance in the second case. We transfer all the weights of the RoBERTa-base model, leaving out the binary classifier layer when fine-tuning to the cognitive dissonance task.
The results of fine-tuning on one or both best transfer model was picked as the model having trained on PDTB and then further fine-tuned on the Debate task as well, as shown in Table \ref{tab:kialo_pdtb}. 
%The metrics reported are based on evaluation on a small annotated dataset of 1714 examples with 68 (3.96\%) dissonant examples.

\subsubsection{AL strategies}
\label{subsec:al-strategies}
Since our annotation process brought about only a small incremental improvement for performance on the rare class yet contributed much to modeling the dominant classes, we hypothesized that using the probability of the rare class as an acquisition strategy in active learning could work just as well as other strategies that are based on diversity and uncertainty sampling. We run our analyses over four other common acquisition strategies by picking out the top 10\% (300 out of an unannotated data pool containing 3,000 examples). We limit to four strategies because of annotation costs and limited time. 
%\sj{We have both \probrareclass{} and PRC as shortened name of probability-of-rare-class. Maybe just use one? I like PRC} \va{Good point: changed to PRC}
\paragraph{\probrareclass{}} 
For a rare, hard class, we use a binary classifier that outputs the probability of rare class learned from the samples encountered so far. This is a computationally inexpensive and simple method that could be easily surpassed by other complex AL strategies but was surprisingly found to be the most effective in this study. The examples from the data pool that are predicted to have the highest probability of rare class by the classification model from previous iteration are selected.

\paragraph{\random{}} 
As a baseline, we randomly sample examples from the data pool, which reflects the natural distribution of classes. Random method has been considered to be a solid baseline to compare against, as many AL strategies do not merit when the annotation pool scales up and collective outliers are missed, as explained in \S\ref{sec:related}.

\paragraph{\entropy{}} We use predictive entropy as the uncertainty-based sampling baseline to compare against. While Least Confident Class (LCC) is a popular strategy to capture samples based on uncertainty, it is calculated based on only one class, working best for binary classification and provides merit within balanced classes, whereas predictive entropy is a generalized form of LCC, and a more popular variant \citep{freeman1965elementary}.

\begin{table}
\centering
\begin{small}
\begin{tabular}{p{0.12\linewidth}p{0.1\linewidth}p{0.1\linewidth}p{0.1\linewidth}p{0.1\linewidth}p{0.1\linewidth}}

\toprule
 & Random & Entropy & CoreSet & \textsc{Cal} & \probrareclass{} \\
\hline
{\color[HTML]{000000} Random}  & \multicolumn{1}{c}{\cellcolor[HTML]{FFFFFF}{\color[HTML]{000000} ×}} & \cellcolor[HTML]{D8E9D1}{\color[HTML]{330001} 12.15\%} & \cellcolor[HTML]{D9EAD2}{\color[HTML]{330001} 11.52\%} & \cellcolor[HTML]{D9EAD3}{\color[HTML]{330001} 10.83\%} & \cellcolor[HTML]{D9EAD3}{\color[HTML]{330001} 11.02\%} \\
{\color[HTML]{000000} Entropy} &   & \multicolumn{1}{c}{\color[HTML]{000000} ×} & \cellcolor[HTML]{8CBC77}{\color[HTML]{330001} 64.68\%} & \cellcolor[HTML]{7BB263}{\color[HTML]{000000} 76.33\%} & \cellcolor[HTML]{6AA84F}{\color[HTML]{000000} 87.98\%} \\
{\color[HTML]{000000} CoreSet} &  & {\color[HTML]{000000} }   & \multicolumn{1}{c}{\color[HTML]{000000} ×}  & \cellcolor[HTML]{95C282}{\color[HTML]{343434} 58.67\%} & \cellcolor[HTML]{90BF7D}{\color[HTML]{330001} 61.65\%} \\
{\color[HTML]{000000} CAL}  & & {\color[HTML]{EFEFEF} } & {\color[HTML]{EFEFEF} } & \multicolumn{1}{c}{\color[HTML]{000000} ×}         & \cellcolor[HTML]{72AD58}{\color[HTML]{000000} 82.98\%}
 \\ 
\hline

\end{tabular}
\caption{\% overlap in the samples picked out by the base model for the five strategies described in \ref{subsec:al-strategies}. Probability of rare class (\textsc{PRC}) has a significant overlap with a state-of-the-art approaches, implying that for the rare class problem, \textsc{PRC} is a computationally inexpensive, alternative acquisition approach. %\has{is the PRC and entropy result correct? 87\% seems quite high for the difference in performance to be so high.}
}
\label{tab:sample-overlap}
\end{small}
\end{table}

\paragraph{\CAL{}}
Contrastive Active Learning \citep{margatina-etal-2021-active} is a state-of-the-art approach that chooses data points that are closely located in the model feature space yet predicted by models to have maximally different likelihoods from each other.
This method is relevant to the task at hand because in rare class problems, it is often difficult for a model to learn the decision boundary around the rare class due to the low number of such samples. Thus we focus on a method that tries to pick out samples at the decision boundary of the rare class.

\paragraph{\coreset{}}
An acquisition method that has worked well as a diversity sampling method is CoreSet \citep{sener2018active}. This method uses a greedy strategy to sample a subset of data that is most representative of the real dataset, i.e. the larger data pool that we sample from.

\subsubsection{Model Update}
To the best of our knowledge, the question of model update in an AL loop has not been explored. We explore two fine-tuning approaches to update the model following annotation of new samples in each round of the active learning loop -- cumulative (\cumulative{}) and iterative (\iterative{}). Figure \ref{fig:AL_loop} provides a visual explanation of the two approaches. 

\paragraph{Cumulative (\cumulative)} 
At each round of the AL loop, the 300 newly annotated samples are combined with the previous ones as the input to fine-tune the classification model from a base pretrained language model.

\paragraph{Iterative  (\iterative{})} 

At each round of the AL loop, the 300 newly annotated samples are used to further fine-tune the model trained during the previous loop.

%\subsection{Model}

\section{Results}

\begin{figure}
    \centering
    \includegraphics[width=7.5cm]{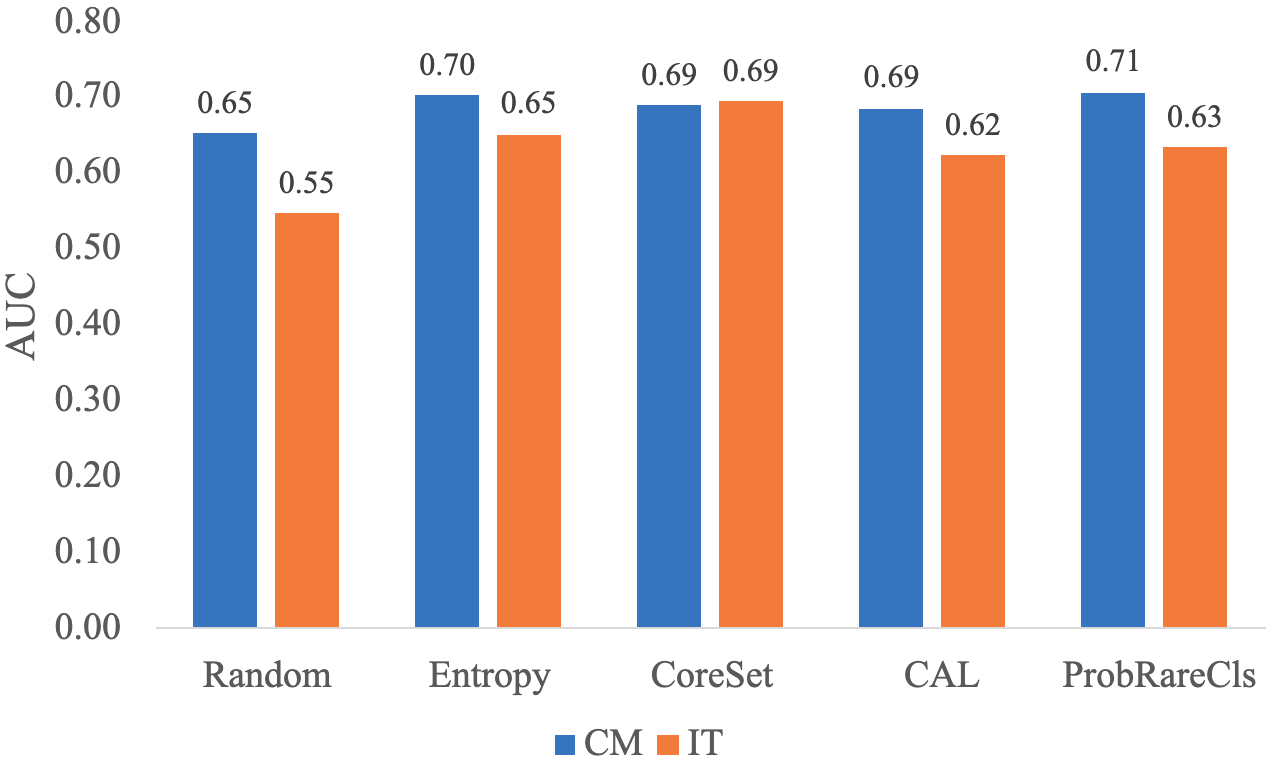} 
    \caption{AUC for the five strategies for \iterative{} and \cumulative{} model updates. This shows that the \cumulative{} model update always performs equally with or better than the \iterative{} update. 
    %This is likely because the \iterative{} approach biases the model to the data distribution of the most recently acquired annotations, while \cumulative{} approach learns on all the data collected so far from scratch, and doesn't get biased to recent data updates. 
}
   
    \label{fig:comparison_metrics}
\end{figure}

\begin{table*}
\centering
\begin{small}
\begin{tabular}{llllll|lllll}
\toprule
        & \multicolumn{5}{c|}{IT}                                                                                             & \multicolumn{5}{c}{CM}                                                                                            \\\midrule
 Selection Strategy & \multicolumn{1}{c}{F1 macro} & \multicolumn{1}{c}{F1 dis} & \multicolumn{1}{c}{Prec dis} & \multicolumn{1}{c}{Rec dis}  & \multicolumn{1}{c|}{AUC} & \multicolumn{1}{c}{F1 macro} & \multicolumn{1}{c}{F1 dis} & \multicolumn{1}{c}{Prec dis} & \multicolumn{1}{c}{Rec dis}  & \multicolumn{1}{c}{AUC} \\ \midrule
Random & 0.556 & 0.175 & 0.119 & \textbf{0.336} & 0.546 & 0.640 &0.362 & 0.397 &  0.334 & 0.652 \\
Entropy & 0.632 & 0.351 & 0.401 & 0.318 & 0.650 & 0.649 & \textbf{0.398} & 0.540 & 0.315 & 0.702 \\
CoreSet & \textbf{0.652} & 0.397 & 0.513 & 0.329 & 0.694 & 0.635 & 0.375 & 0.523 & 0.292 & 0.688 \\
\CAL{} & 0.612 & 0.306 & 0.331 & 0.321 & 0.623 & 0.644 & 0.383 & 0.497 & 0.313 & 0.685 \\
\probrareclass{} & 0.616 & 0.322 & 0.371 & 0.309 & 0.633 & 0.633 &0.382 & \textbf{0.580} & 0.285 & \textbf{0.706} \\
\bottomrule
\end{tabular}

\caption{Comparison of five annotation strategies for iterative (\iterative{}) and cumulative (\cumulative{}) approaches for 2 class classification. 
The metrics are averaged over two iterations of active learning, with 300 new examples annotated in each iteration (adds between 3-10\% samples of dissonance in each round, depending on the strategy). 
\textbf{Bold} represents the best for each reported metric.
The performance of \cumulative{} approach exceeds that of \iterative{} across most acquisition functions, which contrasts with the case of transfer learning step where combining the datasets into one did not help the model as much. 
%[This is likely because the model overfits to the most recently acquired data in \iterative{} approach while could help in generalizing to new domains during transfer learning, but might not add a lot of value when data is collected in the same domain in each iteration of the AL loop.] <-- \sj{assumption}
While the performance on adding 10 to 30 samples of dissonance is not expected to cause large jumps in performance, note that using the \probrareclass{} strategy leads to significant gain in performance in detecting the dissonance class compared to the transfer models from Table \ref{tab:kialo_pdtb_finetuning}. 
}
% We also find that the ProbDis strategy performs the best in terms of the AUC of the model, while Entropy performs the best in terms of F1 dissonance. Note that the samples selected by the two best strategies have ~90\% overlap as shown in Fig \ref{tab:sample-overlap}. 

\label{tab:comparison-AL}
\end{small}

\end{table*}

\subsection{Transfer Learning Models for Cognitive Dissonance}
% The transfer learning models for bootstrapping the active annotation as displayed in 
%\sj{
Table \ref{tab:kialo_pdtb} displays the evaluation of the transfer learning models on bootstrapping the active annotation, revealing that pretraining the large language models on relevant tasks that are specifically designed to mimic the task at hand can lead to better performance.
%}
% Table \ref{tab:kialo_pdtb} shows the ability of the large language models to get better at tasks by pretraining on relevant tasks that are specifically designed to mimic the task at hand. 
% Table \ref{tab:kialo_pdtb} shows that
In addition, the transfer from both Debate and \pdtb{} tasks leads to better results than training the RoBERTa-base model on the Dissonance dataset directly. We also combine the two datasets used in Debate and \pdtb{} and train them at the same time -- similar to the \cumulative{} approach -- to find that Deb;\pdtb{} model still performs better than the model directly trained on the Dissonance dataset. This shows the incredible zero-shot abilities of transfer models for this task.
%shows the metrics obtained after repeating the validation for the collected dataset with a much bigger sample size -- 2,940 examples with 303 dissonant pairs. 
%Comparing tables \ref{tab:kialo_pdtb} and \ref{tab:kialo_pdtb_}, we find that increasing the number of data and \sj{what does this mean?} [improving the dissonance samples] from around 4\% to 10\% through the annotation process significantly improves the metrics for all the candidate transfer learning models.

Furthermore, we explore if continuing to pretrain on a different task after already having pretrained on Debate or \pdtb{} makes a difference. In such case, order of pretraining tasks matters, and there is a much larger gain in the zero-shot performance for \pdtb{}$\rightarrow$Debate compared to Debate$\rightarrow$\pdtb{} as seen in Table \ref{tab:kialo_pdtb_finetuning}. 
When any of these transfer models is further fine-tuned on the dissonance dataset, we find an initial drop in performance. This is explained with the effect of the heterogeneous domain transfer and the small dataset in the $iter_0$ train set. As later shown in Table \ref{tab:comparison-AL}, the performance improves when more samples are collected in the AL iterations. The domain transfer from both tasks (or a combination of them) gives the active annotation a head-start for initial sample selection.

\subsection{Acquisition Strategies}

%\begin{figure}[!ht]
%    \centering
%    \includegraphics[width=7.5cm]{figures/AL_strategies.png} % link to drawing: https://docs.google.com/drawings/d/1whza7Hmtv16NnDHjdIAVJuMDsOVDkQ_OHqo4ADasIKM/edit?usp=sharing
%    \caption{Results of Annotation cost evaluation. The table compares the time taken on an average to annotate each sample picked by the strategies, and the average difficulty. Dissonance class has the difficult samples and this is seen in the high values of timing and difficulty across the both the \probrareclass{} strategies. \bluetodo{with co-authors :this figure vs table 4 -  pick one-- they are the same} }
   
%    \label{fig:AL_comp}
%\end{figure}

Table \ref{tab:sample-overlap} shows the overlap of samples picked out in each iteration from the same larger data pool for the model at iteration 0 (base model). \random{} has the lowest overlaps with all the other strategies. We also find that there is a significant overlap (> $80\%$) in the samples between \entropy{} or \CAL{}, the state-of-the-art approach, and \probrareclass{}. \CAL{} has a higher overlap with \entropy{} rather than \coreset{}, showing that samples deemed to be both highly informative and contrastive by the model are also usually likely to be dissonant. This is contrastive to the prior literature revealing that poor calibration of large language models often renders the models to rarely be uncertain of their outcomes \cite{guo2017calibration}.
All strategies except \random{} have > $55\%$ overlap with each other. This implies that diversity- and uncertainty-based methods are not as different from each other as they theoretically are and inclined to pick similar samples -- hinting that a lot of diversity-based sampling measures mostly pick highly informative samples as well. 
%\sj{
Furthermore, \probrareclass{} tends to choose samples that the ``state-of-the-art'' model also picks in rare-case scenarios, indicating that it could be a computationally inexpensive alternative.
%}
% and in case of rare class scenarios, \probrareclass{} works quite well at picking the samples that a ``state-of-the-art'' model picks.

Table \ref{tab:comparison-AL} shows the results averaged over two rounds of active annotation and learning for five strategies with two types of model updates. While the performance for dissonance class across all strategies do not seem to boost much in a single round of active learning (since adding 300 new annotations adds only between 10-30 dissonance examples in each round), Figure \ref{fig:comparison_metrics} shows that the \cumulative{} approach always performs better than \iterative{}. 
\iterative{} could help models generalize to new domains during transfer learning, but it may not add a lot of value when data is collected in the same domain in each iteration of the AL loop. 
This could be because \iterative{} biases the model towards the distribution of the latest sample set while \cumulative{} implicitly balances all batches of data. 
%This is likely because the model overfits to the most recently acquired data in \iterative{} approach while 

The performance of \random{}-\cumulative{} strategy lags behind the rest of the \cumulative{} strategies. The other strategies perform better than \random{} but one strategy does not offer significant advantages over another, further confirming the observation from Table \ref{tab:sample-overlap} that the AL strategies have a significant overlap and could be choosing very similar samples.

\subsection{Qualitative Evaluation of Annotation Costs}

Table \ref{tab:timing-diff} displays the results of a study on the quality of annotation, measuring subjective difficulty and time taken. We sampled 300 examples from a data pool of 3,000 unannotated examples for each strategy so that the experiment is consistent with the unlabeled pool size used across other experiments for each of the strategies. Of these 300, we picked 125 (from each strategy) to get annotated for their difficulty on a scale of 0-5. This number was chosen based on balancing having enough examples per strategy for meaningful statistics while not taking too much of annotator's time and effort. The annotations were conducted on a simple annotation app that records the time taken to produce the first label an annotator decides on (i.e. any corrections to the label wouldn't count towards the time calculation). The Pearson correlation between the average time taken and the average difficulty value was 0.41.  

%The values within the brackets are the standard deviations
%\todo{Would drop cohen's k from table 3 (mention in text about annotation -- that is where it's first expected) and would change subj. diff to be zscores then use mean of those}

Annotation cost (in terms of time taken to annotate) is known to increase when employing active learning strategies compared to that of a random baseline \cite{settles2008active}. 
We find that \probrareclass{} picks out the ``most difficult'' samples, and takes almost a second longer to annotate than average (average time taken: 12.59s), followed by \entropy{} and \coreset{} strategies -- this complies with \entropy{} picking the most uncertain samples and \coreset{} executing diversity sampling and representing the data better, thus increasing the number of dissonance samples. 
The subjective difficulty reported is the average z-score of difficulty scores picked by the annotators. 
This is done to normalize the variability of subjective ratings.
The inter-rater reliability for the entire exercise was measured using the Cohen's $\kappa$ for two annotators, which was calculated to be  0.37 (fair agreement), with an overlap of 66\%.

%We also find the difficulty of dissonance samples reflected in the Cohen's kappa for 2 annotators -- the strategies with higher percentage of dissonance (\entropy{} and \probrareclass{}) has a low Cohen's kappa, while  the other three strategies have a moderate value of Cohen's kappa. 

\begin{table}[ht]
\centering
\begin{small}

\begin{tabular}{clll}

% cohen's kappa, time, rare class, subjective difficult
\toprule
%& \multicolumn{1}{l}{Cohen's \kappa} 
& \multicolumn{1}{l}{Rare \%} & \multicolumn{1}{l}{Time (s)} & \multicolumn{1}{l}{Subj. diff.}\\\midrule
\random{} 
%& \multicolumn{1}{c}{\cellcolor[HTML]{CEE3C5}{\color[HTML]{000000} 0.30}} 
&  \multicolumn{1}{c}{\cellcolor[HTML]{FFFFFF}{\color[HTML]{000000} 3.20}  }        &  \multicolumn{1}{c}{\cellcolor[HTML]{F6F9FE}{\color[HTML]{000000} 11.96}}
&  \multicolumn{1}{c}
%{\cellcolor[HTML]{F9F2F6}{\color[HTML]{000000} 1.28}} \\
{\cellcolor[HTML]{D9EAD3} {\color[HTML]{000000} -0.065}} \\
\entropy{} 
% &  \multicolumn{1}{c}{\cellcolor[HTML]{FFFFFF}{\color[HTML]{000000} 0.25}} 
&  \multicolumn{1}{c}{\cellcolor[HTML]{EBA55D}{\color[HTML]{000000} 6.80} } &
  \multicolumn{1}{c}{\cellcolor[HTML]{96B7EA}{\color[HTML]{000000} 12.78}} &
   \multicolumn{1}{c}{
   %\cellcolor[HTML]{D5A4BE}{\color[HTML]{000000} 1.35}} \\
   \cellcolor[HTML]{88BA72}{\color[HTML]{000000} 0.035}} \\
\coreset{} 
%&  \multicolumn{1}{c}{\cellcolor[HTML]{6AA84F}{\color[HTML]{000000} \textbf{0.40}}} 
&  \multicolumn{1}{c}{\cellcolor[HTML]{F0B981}{\color[HTML]{000000} 6.00}}&
  \multicolumn{1}{c}{\cellcolor[HTML]{FEFEFF}{\color[HTML]{000000} 11.89}} & 
   \multicolumn{1}{c}{
   %\cellcolor[HTML]{D3A0BB}{\color[HTML]{000000} 1.35} }\\
   \cellcolor[HTML]{85B86F}{\color[HTML]{000000} 0.039} }\\ 
\CAL{}     
%&  \multicolumn{1}{c}{\cellcolor[HTML]{9CC68A}{\color[HTML]{000000} 0.35}} 
& \multicolumn{1}{c}{\cellcolor[HTML]{F6D7B7}{\color[HTML]{000000} 4.80}}&
  \multicolumn{1}{c}{\cellcolor[HTML]{FFFFFF}{\color[HTML]{000000} 11.88}} &
   \multicolumn{1}{c}{
   %\cellcolor[HTML]{FFFFFF}{\color[HTML]{000000} 1.27}} \\
   \cellcolor[HTML]{C9E1C0}{\color[HTML]{000000}-0.045}} \\
PRC 
%&  \multicolumn{1}{c}{\cellcolor[HTML]{ECF4E8}{\color[HTML]{000000} 0.27}} 
&  \multicolumn{1}{c}{\cellcolor[HTML]{E69138}{\color[HTML]{000000} \textbf{7.60}}} &  \multicolumn{1}{c}{\cellcolor[HTML]{3C78D8}{\color[HTML]{000000} \textbf{13.55}}} &  \multicolumn{1}{c}{
%\cellcolor[HTML]{C27BA0}{\color[HTML]{000000} \textbf{1.38}}}\\
\cellcolor[HTML]{6AA84F}{\color[HTML]{000000}\textbf{0.071}}}\\ 

%\midrule 
%All & & 12.47 (9.76) & 1.31 (0.68) & \multicolumn{1}{c}{4.60} \\
\bottomrule
\end{tabular}

\caption{
Evaluation of annotation difficulty by selection strategy. Rare \% is how much the rare class (dissonance) was selected; Time is per instance and subj diff is z-scored subjective rating of difficulty. Our PRC approach selects the most rare class instances but also results in more costly annotations in terms of time and subjective ratings.
}
% Results of Annotation cost evaluation. The table compares the time taken on an average to annotate each sample picked by the strategies, and the average difficulty. Dissonance class has the difficult samples and this is seen in the high values of timing and difficulty across the both the \probrareclass{} strategies. The Cohen's $\kappa$ is reported for three class annotation according to Fig \ref{fig:dissonance_instructions}. This shows that while PRC selects more examples of dissonance, this rare class is also harder to annotate given subjective annotator instructions, and the inter-annotator agreement is lower. On the other hand, CoreSet has the highest $\kappa$ (moderate agreement) with median difficulty but compromises on the \% dissonance samples. 

\label{tab:timing-diff}
\end{small}
\end{table}

\begin{table}[ht]
\begin{small}
\begin{tabular}{lp{0.1\linewidth}llll}
\toprule
& F1 macro & F1 dis & Prec & Rec & AUC \\
\midrule
model$_{iter0}$ & 0.623 & 0.332 & 0.364 & 0.306 & 0.634 \\
\midrule
Sm Deb & 0.667 & 0.419 & 0.510  & \textbf{0.355} & 0.702 \\
Sm \pdtb{}$\rightarrow$Deb  & 0.658 & 0.389 & 0.483  & 0.327 & 0.707 \\
Big Deb & 0.647      & 0.417    & \textbf{0.695}  & 0.298 & \textbf{0.753} \\
Big \pdtb{}$\rightarrow$Deb  & \textbf{0.669} & \textbf{0.425} & 0.536  & 0.352 & 0.711\\
\bottomrule
\end{tabular}
\caption{The final dataset tested on the best transfer models from Tables \ref{tab:kialo_pdtb} and \ref{tab:kialo_pdtb_finetuning} with \cumulative{} approach.  These models could subsequently be used to obtain newer samples more efficiently. model$_{iter0}$ refers to the best model from continued training on Table \ref{tab:kialo_pdtb_finetuning}, with scores reported on the final test and dev sets.}
\label{tab:final-dataset}
\end{small}
\end{table}
% \has{Good to show a baseline at the top of original iter0 results before all of the extra data collection. }
%Hopefully, this performs the best. 
% Based on the performance on the bigger dataset, we find that more the number of dissonance samples to learn from, the better the model performs.

\subsection{A final dataset: Putting it all together.}
%\sj{
We release two versions of train data: small and big; along with the development and test data (see \S\ref{subsec:anno_dataset})
The \textit{small} set comprises the 2,924 examples which were use for the active learning experiments discussed previously.
Building on our learnings from the active learning experiments, we created a second (\textit{big}) data set with 6,649 examples that includes the small plus an additional 3,725 examples  derived over more rounds of active learning restricted to the \probrareclass{} or \entropy{} strategies.   
It contains 692 dissonant samples, comprising 
 10.40\% among all. 

Table \ref{tab:final-dataset} reports the performance improvement from using this final larger dataset, yielding the best performance so far with AUC > 0.75.
% results revealing that the more number of dissonant samples to learn from, the better the model performs.
%}

% We also add additional 2,524 annotations with 329 dissonance samples collected pover 8 more rounds of active annotation with a combination of \probrareclass{} and \entropy{} strategies, and this totals to 513 diss out of 5,440 total samples - much bigger dataset with 9.41\% dissonance samples as training.  We report the results as Big in Table 6.
%

%A small discussion of Table \ref{tab:final-dataset}.

%\bluetodo{Check with co-authors: If we combine the old dataset collected during summer -- the result is so much better if we do (see table 6)! A: good to include; AL is step toward a full dataset and this is what we have now ended up with to train a model -- it should be a benchmark for others using the data to beat.} 

\section{Conclusion}
%In this work  In order to fo comprised of comparison of five common acquisition strategies for active learning for the rare class problem,  find <x> \bluetodo{fill x} to perform the best; (ii) defining and comparing two different approaches -- cumulative and iterative model updates in the AL literature and finding the cumulative update to work better, (iii) qualitatively comparing the active annotation strategies and finding <y> \bluetodo{fill y} to be the best, (iv) a novel dataset for identifying cognitive dissonance on social media

In this work, we have systematically studied approaches to key steps of active learning for tackling a rare-class modeling using a modern large language-modeling approach. 
While transformer-based systems have enabled greater accuracies with fewer training examples, data acquisition obstacles still persist for rare-class tasks -- when the class label is very infrequent (e.g. < 5\% of samples).
We examined pool-based active annotation and learning in a real-world, rare class, natural language setting by exploring five common acquisition strategies with two different model update approaches.
We found that a relatively simple acquisition using the probability of rare class for a model could lead to significant improvement in the rare class samples.
We also qualitatively analyzed the data samples extracted from each data acquisition strategy by using subjective scoring and timing the annotators, finding \probrareclass{} to be the most difficult to annotate, while also remaining the best method to improve rare class samples and model performance.
Our final dataset of approximately 6,000 examples is made available along with an implementation of the PRC method and our state-of-the-art model for cognitive dissonance detection.

\section{Limitations}
%We do not go into finding the best model architecture for this task, and instead focus on data-centric improvements.  
%not sure I'd consider active learning as simply data-centric

%\va{
We use RoBERTa-base models trained on a single 12GB memory GPU (we used a NVIDIA Titan XP graphics card) for our experiments. 
Obtaining annotations for cognitive dissonance are limited by the availability of annotators and is not easily scalable in crowdsourcing platforms due to the required training and expertise in identifying dissonance. Due to this limitation, only two iterations of the AL loop for each setting were feasible for experiments.
The transfer learning experiments in this paper were limited to two similar tasks, but there might be other tasks that could further improve or exceed the zero-shot performance of the models to cold start the active learning. 

We focus on fine-tuning and active learning selection strategies to improve performance of rare-class classification for a specific task: dissonance detection across discourse units. 
Therefore, further work would be necessary to determine if the findings extend to other tasks. 
The performance of the neural parser on splitting tweets into discourse units can produce parses that are imperfect but the annotators and our systems worked off it's output regardless to keep the process consistent. 
An improved discourse parser may also lead to improved annotator agreement and/or classifier accuracy.

\section{Ethics Statement}
%\va{

%attempting to influence beliefs and behaviors without users' awareness. 
The dataset for annotation was created from public social media posts with all usernames, phone numbers, addresses, and URLs removed.
%using langid.py \citep{lui-baldwin-2012-langid}.
The research was approved by an academic institutional ethics review board. 
All of our work was restricted to document-level information; no user-level information was used. According to Twitter User Agreement, no further user content is required to use the publicly available data.
 
%}

%\section*{Acknowledgements}

%\sj{https://aclrollingreview.org/responsibleNLPresearch/}

% scope of your claims, e.g., if you only tested your approach on a few datasets, languages, or did a few runs.
%\sj{Our work of comparing active acquisition strategies scopes one dataset that consists of English language only, thus we acknowledge that the results may be different in various circumstances.} %, i.e. class distributions, class types, languages, etc.}

Our work of comparing active acquisition strategies scopes one dataset that consists of English language only. 
The results may be different for other languages or time intervals. %It also might not be fully representative of the general population due to any existing biases among Twitter users that produce English tweets.

% potential risks
%All the documents were anonymized.
%Cognitive dissonance detection could be used by adversaries to find vulnerable people to influence their behaviors.
%borrowed from kialo paper
The detection of dissonance has many beneficial applications such as understanding belief trends and study of mental health from consenting individuals.
However, it also could be used toward manipulative goals via targeted messaging to influence beliefs potential without users' awareness of such goals, a use-case that this work does not intend.
Further, while we hope such models could be used to help better understand and assess mental health, clinical evaluations would need to be conducted before our models are integrated into any mental health practice. 
Additionally, the associations found should be taken in the context of study sample, which was limited to English language posts from Twitter which tends to be younger on average than the general population. 

%\sj{The dataset that we release from this paper, which contains [strong] \va{can we say they "strong" labels of mental states? they're just labeling some linguistic expressions of mental states?} labels regarding mental states, was constructed using criteria that may not be fully objective. Therefore, we suggest not to use our dataset and models to determine a person's mental condition or for diagnosis and other clinical purposes.}

The dataset that we release from this paper, which contains labels of expressions of some mental states, was constructed using criteria that may not be fully objective. Therefore, we suggest not to use our dataset and models to determine a person's mental condition or for diagnosis and other clinical purposes.

\section*{Acknowledgements}
We thank Lucie Flek (CAISA lab, Philipps Universität Marburg) and Ji-Ung Lee (UKP Lab, TU Darmstadt) for insightful discussions about this work.

This work was supported by DARPA via Young Faculty Award grant \#W911NF-20-1-0306 to H. Andrew Schwartz at Stony Brook University; the conclusions and opinions expressed are attributable only to the authors and should not be construed as those of DARPA or the U.S. Department of Defense. 
\bibliography{anthology,custom}

\begin{thebibliography}{45}
\expandafter\ifx\csname natexlab\endcsname\relax\def\natexlab#1{#1}\fi

\bibitem[{Al-Stouhi and Reddy(2016)}]{al2016transfer}
Samir Al-Stouhi and Chandan~K Reddy. 2016.
\newblock Transfer learning for class imbalance problems with inadequate data.
\newblock \emph{Knowledge and information systems}, 48(1):201--228.

\bibitem[{Ash et~al.(2019)Ash, Zhang, Krishnamurthy, Langford, and
  Agarwal}]{badge}
Jordan~T Ash, Chicheng Zhang, Akshay Krishnamurthy, John Langford, and Alekh
  Agarwal. 2019.
\newblock Deep batch active learning by diverse, uncertain gradient lower
  bounds.
\newblock \emph{arXiv preprint arXiv:1906.03671}.

\bibitem[{Chapman et~al.(1993)Chapman, Wong, and Smith}]{chapman1993self}
Simon Chapman, Wai~Leng Wong, and Wayne Smith. 1993.
\newblock Self-exempting beliefs about smoking and health: differences between
  smokers and ex-smokers.
\newblock \emph{American journal of public health}, 83(2):215--219.

\bibitem[{Choi et~al.(2021)Choi, Yi, Kim, Choo, Kim, Chang, Gwon, and
  Chang}]{Choi_2021_CVPR}
Jongwon Choi, Kwang~Moo Yi, Jihoon Kim, Jinho Choo, Byoungjip Kim, Jinyeop
  Chang, Youngjune Gwon, and Hyung~Jin Chang. 2021.
\newblock Vab-al: Incorporating class imbalance and difficulty with variational
  bayes for active learning.
\newblock In \emph{Proceedings of the IEEE/CVF Conference on Computer Vision
  and Pattern Recognition (CVPR)}, pages 6749--6758.

\bibitem[{Citovsky et~al.(2021)Citovsky, DeSalvo, Gentile, Karydas,
  Rajagopalan, Rostamizadeh, and Kumar}]{cluster_margin}
Gui Citovsky, Giulia DeSalvo, Claudio Gentile, Lazaros Karydas, Anand
  Rajagopalan, Afshin Rostamizadeh, and Sanjiv Kumar. 2021.
\newblock \href
  {https://proceedings.neurips.cc/paper/2021/file/64254db8396e404d9223914a0bd355d2-Paper.pdf}
  {Batch active learning at scale}.
\newblock In \emph{Advances in Neural Information Processing Systems},
  volume~34, pages 11933--11944. Curran Associates, Inc.

\bibitem[{Dalgaard-Nielsen(2013)}]{dalgaard2013promoting}
Anja Dalgaard-Nielsen. 2013.
\newblock Promoting exit from violent extremism: Themes and approaches.
\newblock \emph{Studies in Conflict \& Terrorism}, 36(2):99--115.

\bibitem[{Day and Khoshgoftaar(2017)}]{day2017survey}
Oscar Day and Taghi~M Khoshgoftaar. 2017.
\newblock A survey on heterogeneous transfer learning.
\newblock \emph{Journal of Big Data}, 4(1):1--42.

\bibitem[{Ducoffe and Precioso(2018)}]{ducoffe2018adversarial}
Melanie Ducoffe and Frederic Precioso. 2018.
\newblock Adversarial active learning for deep networks: a margin based
  approach.
\newblock \emph{arXiv preprint arXiv:1802.09841}.

\bibitem[{Freeman(1965)}]{freeman1965elementary}
Linton~C Freeman. 1965.
\newblock \emph{Elementary applied statistics: for students in behavioral
  science}.
\newblock New York: Wiley.

\bibitem[{Gissin and Shalev-Shwartz(2019)}]{gissin2019discriminative}
Daniel Gissin and Shai Shalev-Shwartz. 2019.
\newblock Discriminative active learning.
\newblock \emph{arXiv preprint arXiv:1907.06347}.

\bibitem[{Guo et~al.(2017)Guo, Pleiss, Sun, and
  Weinberger}]{guo2017calibration}
Chuan Guo, Geoff Pleiss, Yu~Sun, and Kilian~Q Weinberger. 2017.
\newblock On calibration of modern neural networks.
\newblock In \emph{International conference on machine learning}, pages
  1321--1330. PMLR.

\bibitem[{Haines and Xiang(2014)}]{haines2014active}
Tom~SF Haines and Tao Xiang. 2014.
\newblock Active rare class discovery and classification using dirichlet
  processes.
\newblock \emph{International Journal of Computer Vision}, 106(3):315--331.

\bibitem[{Harmon-Jones and Harmon-Jones(2007)}]{harmon2007cognitive}
Eddie Harmon-Jones and Cindy Harmon-Jones. 2007.
\newblock Cognitive dissonance theory after 50 years of development.
\newblock \emph{Zeitschrift f{\"u}r Sozialpsychologie}, 38(1):7--16.

\bibitem[{Harmon-Jones et~al.(2008)Harmon-Jones, Harmon-Jones, Fearn, Sigelman,
  and Johnson}]{harmon2008left}
Eddie Harmon-Jones, Cindy Harmon-Jones, Meghan Fearn, Jonathan~D Sigelman, and
  Peter Johnson. 2008.
\newblock Left frontal cortical activation and spreading of alternatives: tests
  of the action-based model of dissonance.
\newblock \emph{Journal of personality and social psychology}, 94(1):1.

\bibitem[{Harmon-Jones and Mills(2019)}]{harmon2019introduction}
Eddie Harmon-Jones and Judson Mills. 2019.
\newblock An introduction to cognitive dissonance theory and an overview of
  current perspectives on the theory.
\newblock \emph{Cognitive dissonance: Reexamining a pivotal theory in
  psychology}.

\bibitem[{Hospedales et~al.(2013)Hospedales, Gong, and
  Xiang}]{rare_class_disco_2011}
Timothy~M. Hospedales, Shaogang Gong, and Tao Xiang. 2013.
\newblock \href {https://doi.org/10.1109/TKDE.2011.231} {Finding rare classes:
  Active learning with generative and discriminative models}.
\newblock \emph{IEEE Transactions on Knowledge and Data Engineering},
  25(2):374--386.

\bibitem[{Houlsby et~al.(2011)Houlsby, Husz{\'a}r, Ghahramani, and
  Lengyel}]{houlsby2011bayesian}
Neil Houlsby, Ferenc Husz{\'a}r, Zoubin Ghahramani, and M{\'a}t{\'e} Lengyel.
  2011.
\newblock Bayesian active learning for classification and preference learning.
\newblock \emph{arXiv preprint arXiv:1112.5745}.

\bibitem[{Kale and Liu(2013)}]{kale2013accelerating}
David Kale and Yan Liu. 2013.
\newblock Accelerating active learning with transfer learning.
\newblock In \emph{2013 IEEE 13th International Conference on Data Mining},
  pages 1085--1090. IEEE.

\bibitem[{Karamcheti et~al.(2021)Karamcheti, Krishna, Fei-Fei, and
  Manning}]{karamcheti-etal-2021-mind}
Siddharth Karamcheti, Ranjay Krishna, Li~Fei-Fei, and Christopher Manning.
  2021.
\newblock \href {https://doi.org/10.18653/v1/2021.acl-long.564} {Mind your
  outliers! investigating the negative impact of outliers on active learning
  for visual question answering}.
\newblock In \emph{Proceedings of the 59th Annual Meeting of the Association
  for Computational Linguistics and the 11th International Joint Conference on
  Natural Language Processing (Volume 1: Long Papers)}, pages 7265--7281,
  Online. Association for Computational Linguistics.

\bibitem[{Kothawade et~al.(2021)Kothawade, Beck, Killamsetty, and
  Iyer}]{similar_neurips}
Suraj Kothawade, Nathan Beck, Krishnateja Killamsetty, and Rishabh Iyer. 2021.
\newblock \href
  {https://proceedings.neurips.cc/paper/2021/file/9af08cda54faea9adf40a201794183cf-Paper.pdf}
  {Similar: Submodular information measures based active learning in realistic
  scenarios}.
\newblock In \emph{Advances in Neural Information Processing Systems},
  volume~34, pages 18685--18697. Curran Associates, Inc.

\bibitem[{Liang et~al.(2019)Liang, Li, and Yin}]{liang2019new}
Yichan Liang, Jianheng Li, and Jian Yin. 2019.
\newblock A new multi-choice reading comprehension dataset for curriculum
  learning.
\newblock In \emph{Asian Conference on Machine Learning}, pages 742--757. PMLR.

\bibitem[{Liu et~al.(2019)Liu, Ott, Goyal, Du, Joshi, Chen, Levy, Lewis,
  Zettlemoyer, and Stoyanov}]{liu2019roberta}
Yinhan Liu, Myle Ott, Naman Goyal, Jingfei Du, Mandar Joshi, Danqi Chen, Omer
  Levy, Mike Lewis, Luke Zettlemoyer, and Veselin Stoyanov. 2019.
\newblock Roberta: A robustly optimized bert pretraining approach.
\newblock \emph{arXiv preprint arXiv:1907.11692}.

\bibitem[{Mann and Thompson(1987)}]{mann1987rhetorical}
William~C Mann and Sandra~A Thompson. 1987.
\newblock \emph{Rhetorical structure theory: A theory of text organization}.
\newblock University of Southern California, Information Sciences Institute Los
  Angeles.

\bibitem[{Margatina et~al.(2021)Margatina, Vernikos, Barrault, and
  Aletras}]{margatina-etal-2021-active}
Katerina Margatina, Giorgos Vernikos, Lo{\"\i}c Barrault, and Nikolaos Aletras.
  2021.
\newblock \href {https://doi.org/10.18653/v1/2021.emnlp-main.51} {Active
  learning by acquiring contrastive examples}.
\newblock In \emph{Proceedings of the 2021 Conference on Empirical Methods in
  Natural Language Processing}, pages 650--663, Online and Punta Cana,
  Dominican Republic. Association for Computational Linguistics.

\bibitem[{McGrath(2017)}]{mcgrath2017dealing}
April McGrath. 2017.
\newblock Dealing with dissonance: A review of cognitive dissonance reduction.
\newblock \emph{Social and Personality Psychology Compass}, 11(12):e12362.

\bibitem[{Munjal et~al.(2022)Munjal, Hayat, Hayat, Sourati, and
  Khan}]{Munjal_2022_CVPR}
Prateek Munjal, Nasir Hayat, Munawar Hayat, Jamshid Sourati, and Shadab Khan.
  2022.
\newblock Towards robust and reproducible active learning using neural
  networks.
\newblock In \emph{Proceedings of the IEEE/CVF Conference on Computer Vision
  and Pattern Recognition (CVPR)}, pages 223--232.

\bibitem[{Netzer et~al.(2011)Netzer, Wang, Coates, Bissacco, Wu, and
  Ng}]{marginentropy2011}
Yuval Netzer, Tao Wang, Adam Coates, Alessandro Bissacco, Bo~Wu, and Andrew~Y.
  Ng. 2011.
\newblock \href
  {http://ufldl.stanford.edu/housenumbers/nips2011_housenumbers.pdf} {Reading
  digits in natural images with unsupervised feature learning}.
\newblock In \emph{NIPS Workshop on Deep Learning and Unsupervised Feature
  Learning 2011}.

\bibitem[{Ng(2021)}]{ng2021mlops}
Andrew Ng. 2021.
\newblock Mlops: from model-centric to data-centric ai.
\newblock \emph{Online unter https://www. deeplearning.
  ai/wp-content/uploads/2021/06/MLOps-From-Model-centric-to-Data-centricAI. pdf
  [Zugriffam09. 09.2021] Search in}.

\bibitem[{Polanyi(1988)}]{polanyi1988formal}
Livia Polanyi. 1988.
\newblock A formal model of the structure of discourse.
\newblock \emph{Journal of pragmatics}, 12(5-6):601--638.

\bibitem[{Prasad et~al.(2008)Prasad, Dinesh, Lee, Miltsakaki, Robaldo, Joshi,
  and Webber}]{prasad2008penn}
Rashmi Prasad, Nikhil Dinesh, Alan Lee, Eleni Miltsakaki, Livio Robaldo,
  Aravind Joshi, and Bonnie Webber. 2008.
\newblock The penn discourse treebank 2.0.
\newblock In \emph{Proceedings of the Sixth International Conference on
  Language Resources and Evaluation (LREC'08)}.

\bibitem[{Sener and Savarese(2018)}]{sener2018active}
Ozan Sener and Silvio Savarese. 2018.
\newblock \href {https://openreview.net/forum?id=H1aIuk-RW} {Active learning
  for convolutional neural networks: A core-set approach}.
\newblock In \emph{International Conference on Learning Representations}.

\bibitem[{Settles et~al.(2008)Settles, Craven, and
  Friedland}]{settles2008active}
Burr Settles, Mark Craven, and Lewis Friedland. 2008.
\newblock Active learning with real annotation costs.
\newblock In \emph{Proceedings of the NIPS workshop on cost-sensitive
  learning}, volume~1. Vancouver, CA:.

\bibitem[{Shannon(1948)}]{shannon1948mathematical}
Claude~Elwood Shannon. 1948.
\newblock A mathematical theory of communication.
\newblock \emph{The Bell system technical journal}, 27(3):379--423.

\bibitem[{Shen et~al.(2017)Shen, Yun, Lipton, Kronrod, and
  Anandkumar}]{shen2017deep}
Yanyao Shen, Hyokun Yun, Zachary~C Lipton, Yakov Kronrod, and Animashree
  Anandkumar. 2017.
\newblock Deep active learning for named entity recognition.
\newblock \emph{arXiv preprint arXiv:1707.05928}.

\bibitem[{Son et~al.(2022)Son, Varadarajan, and Schwartz}]{discre2022}
Youngseo Son, Vasudha Varadarajan, and H.~Andrew Schwartz. 2022.
\newblock Discourse relation embeddings: Representing the relations between
  discourse segments in social media.
\newblock In \emph{Proceedings of the Workshop on Unimodal and Multimodal
  Induction of Linguistic Structures (UM-IoS)}. Association for Computational
  Linguistics.

\bibitem[{Taboada and Mann(2006)}]{taboada2006rhetorical}
Maite Taboada and William~C Mann. 2006.
\newblock Rhetorical structure theory: Looking back and moving ahead.
\newblock \emph{Discourse studies}, 8(3):423--459.

\bibitem[{Tsvigun et~al.(2022)Tsvigun, Shelmanov, Kuzmin, Sanochkin, Larionov,
  Gusev, Avetisian, and Zhukov}]{tsvigun-etal-2022-towards}
Akim Tsvigun, Artem Shelmanov, Gleb Kuzmin, Leonid Sanochkin, Daniil Larionov,
  Gleb Gusev, Manvel Avetisian, and Leonid Zhukov. 2022.
\newblock \href {https://doi.org/10.18653/v1/2022.findings-naacl.90} {Towards
  computationally feasible deep active learning}.
\newblock In \emph{Findings of the Association for Computational Linguistics:
  NAACL 2022}, pages 1198--1218, Seattle, United States. Association for
  Computational Linguistics.

\bibitem[{Varadarajan et~al.(2022)Varadarajan, Soni, Wang, Luhmann, Schwartz,
  and Inoue}]{varadarajan2022disso}
Vasudha Varadarajan, Nikita Soni, Weixi Wang, Christian Luhmann, H.~Andrew
  Schwartz, and Naoya Inoue. 2022.
\newblock \href {https://aclanthology.org/2022.nlpcss-1.16} {Detecting
  dissonant stance in social media: The role of topic exposure}.
\newblock In \emph{Proceedings of the Fifth Workshop on Natural Language
  Processing and Computational Social Science (NLP+CSS)}. Association for
  Computational Linguistics.

\bibitem[{Wang and Shang(2014)}]{Wang2014ANA}
Dan Wang and Yi~Shang. 2014.
\newblock A new active labeling method for deep learning.
\newblock \emph{2014 International Joint Conference on Neural Networks
  (IJCNN)}, pages 112--119.

\bibitem[{Wang et~al.(2018)Wang, Li, and Yang}]{wang-etal-2018-toward}
Yizhong Wang, Sujian Li, and Jingfeng Yang. 2018.
\newblock \href {https://doi.org/10.18653/v1/D18-1116} {Toward fast and
  accurate neural discourse segmentation}.
\newblock In \emph{Proceedings of the 2018 Conference on Empirical Methods in
  Natural Language Processing}, pages 962--967, Brussels, Belgium. Association
  for Computational Linguistics.

\bibitem[{Wu et~al.(2022)Wu, Xiao, Sun, Zhang, Ma, and He}]{survey_HITL}
Xingjiao Wu, Luwei Xiao, Yixuan Sun, Junhang Zhang, Tianlong Ma, and Liang He.
  2022.
\newblock \href {https://doi.org/https://doi.org/10.1016/j.future.2022.05.014}
  {A survey of human-in-the-loop for machine learning}.
\newblock \emph{Future Generation Computer Systems}, 135:364--381.

\bibitem[{Yuan et~al.(2020)Yuan, Lin, and Boyd-Graber}]{yuan-etal-2020-cold}
Michelle Yuan, Hsuan-Tien Lin, and Jordan Boyd-Graber. 2020.
\newblock \href {https://doi.org/10.18653/v1/2020.emnlp-main.637} {Cold-start
  active learning through self-supervised language modeling}.
\newblock In \emph{Proceedings of the 2020 Conference on Empirical Methods in
  Natural Language Processing (EMNLP)}, pages 7935--7948, Online. Association
  for Computational Linguistics.

\bibitem[{Zhan et~al.(2022)Zhan, Wang, Huang, Xiong, Dou, and
  Chan}]{zhan2022comparative}
Xueying Zhan, Qingzhong Wang, Kuan-hao Huang, Haoyi Xiong, Dejing Dou, and
  Antoni~B Chan. 2022.
\newblock A comparative survey of deep active learning.
\newblock \emph{arXiv preprint arXiv:2203.13450}.

\bibitem[{Zhang and Plank(2021)}]{cartography}
Mike Zhang and Barbara Plank. 2021.
\newblock Cartography active learning.
\newblock \emph{arXiv preprint arXiv:2109.04282}.

\bibitem[{Zhuang et~al.(2021)Zhuang, Qi, Duan, Xi, Zhu, Zhu, Xiong, and
  He}]{comprehensiveTL}
Fuzhen Zhuang, Zhiyuan Qi, Keyu Duan, Dongbo Xi, Yongchun Zhu, Hengshu Zhu, Hui
  Xiong, and Qing He. 2021.
\newblock \href {https://doi.org/10.1109/JPROC.2020.3004555} {A comprehensive
  survey on transfer learning}.
\newblock \emph{Proceedings of the IEEE}, 109(1):43--76.

\end{thebibliography}
\bibliographystyle{acl_natbib}

%\appendix

%\section{Example Appendix}
%\label{sec:appendix}

%This is an appendix.

\end{document}